%% file: root.tex
\documentclass[letterpaper, 10 pt, conference]{ieeeconf}  

\IEEEoverridecommandlockouts                              

\usepackage{graphicx} 
\usepackage{amsmath} 
\usepackage{amssymb}  
\usepackage{csquotes}
\usepackage{mdframed}
\usepackage{soul}
\usepackage{url}
\usepackage{multirow}
\usepackage{balance}
\usepackage{pgfplots}
\usepackage[export]{adjustbox}
\usepackage{capt-of}
\usepackage{balance}

\makeatletter
\let\NAT@parse\undefined
\makeatother
\usepackage[backref=page]{hyperref}
\usepackage{cleveref}

\pgfplotsset{compat=1.13}

\crefname{section}{Sec.}{Sections}
\Crefname{section}{Sec.}{Sections}
\crefname{figure}{Fig.}{Figs.}
\Crefname{figure}{Fig.}{Figs.}
\crefname{table}{Table}{Table}
\Crefname{table}{Table}{Table}
\crefname{equation}{Eq.}{Eqs.}
\Crefname{equation}{Eq.}{Eqs.}
\crefname{listing}{Listing}{Listings}
\Crefname{listing}{Listing}{Listings}

\input{util}

\title{\LARGE \bf
Episodic Memory Verbalization using \\ Hierarchical Representations of Life-Long Robot Experience
}

\author{Leonard Bärmann$^{1}$, Chad DeChant$^{2}$, Joana Plewnia$^{1}$, Fabian Peller-Konrad$^{1}$,\\ Daniel Bauer$^{2}$, Tamim Asfour$^{1}$ and Alex Waibel$^{1}$
\thanks{\ack}
\thanks{$^{1}$Institute for Anthropomatics and Robotics, 
        Karlsruhe Institute of Technology, Germany
        E-mails: {\tt\small \{baermann,asfour\}@kit.edu}}%
\thanks{$^{2}$Computer Science Department, Columbia University, NY, United States
        E-mail: {\tt\small chad.dechant@columbia.edu}}%
}

\begin{document}

\maketitle
\thispagestyle{empty}
\pagestyle{empty}


\input{content}

\renewcommand*{\backref}[1]{}
\renewcommand*{\backrefalt}[4]{%
  \ifcase #1 %
  \or
    Cited on page #2.%
  \else
    Cited on pages #2.%
  \fi%
}

\balance
\bibliographystyle{IEEEtran}
\bibliography{root}

\end{document}

%% file: util.tex
\usepackage[per-mode=fraction]{siunitx}
\usepackage{comment}
\usepackage[normalem]{ulem}
\usepackage{xspace}
\usepackage{xcolor}

\definecolor{OliveGreen}{RGB}{0,170,0}

\newcommand{\ie}{i.\,e.,\xspace}
\newcommand{\eg}{e.\,g.,\xspace}

\newcommand{\armar}{\mbox{\textsc{Armar}}\xspace}

\newcommand{\armarVII}{\mbox{\textsc{Armar}-7}\xspace}
\newcommand{\armarx}{\mbox{ArmarX}\xspace}

\newcommand{\ack}{This work has been supported by the Baden-Württemberg Ministry of Science, Research and the Arts (MWK) as part of the state's ``digital@bw'' digitization strategy in the context of the Real-World Lab ``Robotics AI'', the Carl Zeiss Foundation through the JuBot project and the German Federal Ministry of Research, Technology and Space (BMFTR) under the Robotics Institute Germany (RIG).}

%% file: content.tex
\begin{abstract}
Verbalization of robot experience, \ie  summarization of and question answering about a robot's past, is a crucial ability for improving human-robot interaction.
Previous works applied rule-based systems or fine-tuned deep models to verbalize short (several-minute-long) streams of episodic data, limiting generalization and transferability.
In our work, we apply large pretrained models to tackle this task with zero or few examples, and specifically focus on verbalizing life-long experiences.
For this, we derive a tree-like data structure from episodic memory (EM), with lower levels representing raw perception and proprioception data, and higher levels abstracting events to natural language concepts.
Given such a hierarchical representation built from the experience stream, we apply a large language model as an agent to interactively search the EM given a user's query, dynamically expanding (initially collapsed) tree nodes to find the relevant information.
The approach keeps computational costs low even when scaling to months of robot experience data.
We evaluate our method on simulated household robot data, human egocentric videos, and real-world robot recordings, demonstrating its flexibility and scalability.
\end{abstract}
\noindent\textbf{\footnotesize Code, data and demo videos at \href{https://hierarchical-emv.github.io}{hierarchical-emv.github.io}.}

\newcommand{\codesize}[0]{\footnotesize}
\newcommand{\paragraphHeading}[1]{\vspace{0.cm}\noindent\textbf{#1:}\ \,}
\newcommand{\ours}[0]{\textsc{H-Emv}\xspace}

\section{Introduction}

\begin{figure}
    \centering
    \includegraphics[width=1\linewidth]{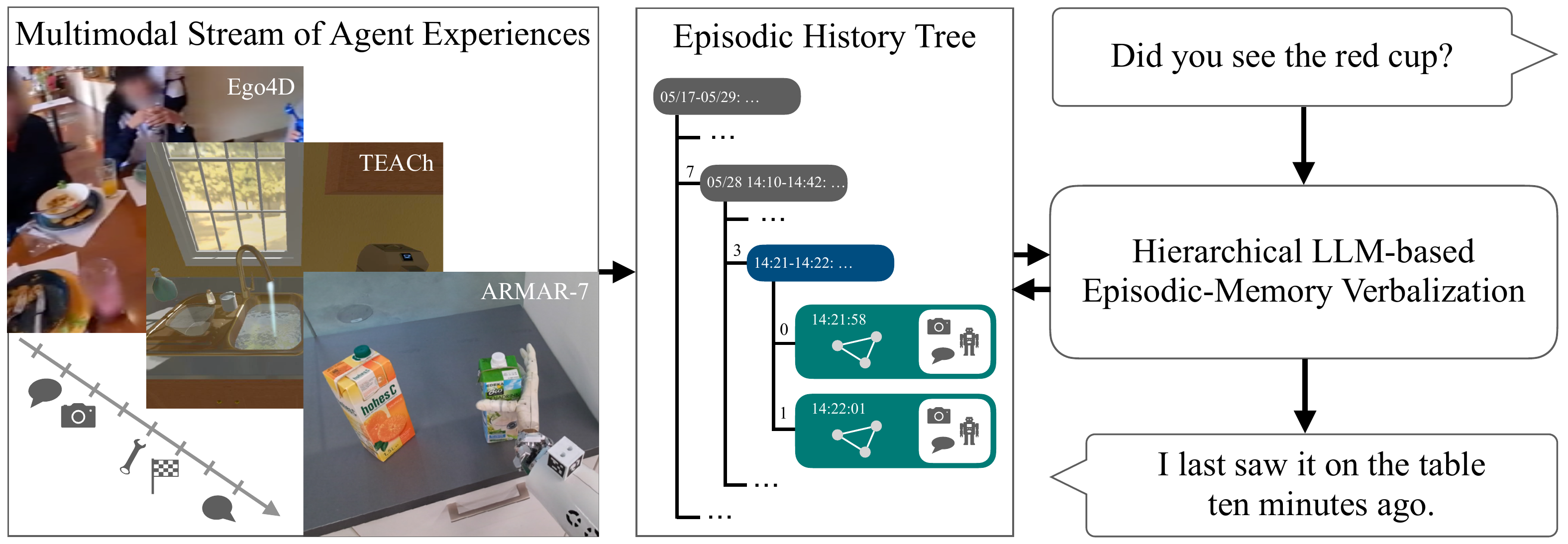}
    \caption{Our system answers queries about life-long experience of an agent (human or robotic) by exploring a tree representation of episodic memory.}
    \label{fig:title}
\end{figure}

Verbalizing their own experiences is an important ability robots should have to improve natural and intuitive human-robot interaction \cite{barmann_deep_2021,rosenthal_verbalization_2016,zhu_autonomous_2017,dechant_toward_2021,katuwandeniya_what_2025}.
It involves summarization of and question answering (QA) about a robot's past actions, observations and interactions, such as the dialog shown on the right of \cref{fig:title}.
Building a representation of an agent's Episodic Memory (EM)~\cite{tulving_episodic_1972} is crucial to enable such verbalizations, as a system must efficiently store the information from the continuous stream of experience, organize it, and retrieve relevant past events from its EM in response to a user's query.
This is particularly challenging as the time horizon of the EM grows.

Existing work on Episodic Memory Verbalization (EMV) either relies on rule-based verbalization of log files \cite{rosenthal_verbalization_2016,zhu_autonomous_2017}, or fine-tuning deep models on hand-crafted or auto-generated datasets \cite{barmann_deep_2021,dechant_learning_2023} to perform QA and summarization tasks given the recorded experiences.
Both approaches are limited, as they require designing vast numbers of rules or collecting large amounts of experience data.

To avoid training a system, which typically entails collecting large amounts of multimodal experience data, previous works~\cite{zeng_socratic_2023,liu_reflect_2023} use language-based representations of the past which can be obtained from pretrained multimodal models.
Given such a language-based representation of an agent's history of episodic events, one practical way to perform QA is to pass the question and the history to a large language model (LLM) and prompt it to produce an answer.
While this works nicely for short histories \cite{zeng_socratic_2023,islam_video_2024}, in this paper, we focus on how to scale such approaches for verbalization of \emph{life-long} experience streams.
Although recent LLMs offer increasingly long context windows (\ie the maximum number of tokens they can process), up to 2M tokens \cite{gemini_team_gemini_2024}, previous studies~\cite{hsieh_ruler_2024,liu_lost_2024} have shown that these models have difficulty in using all information contained in such long contexts.
Furthermore, the computation of transformer models scales quadratically with context length -- reducing the number of tokens is thus time- and cost-effective.

Therefore, to scale EMV to life-long experience streams while maintaining a low token budget, we propose to derive a tree-like representation from EM and use an LLM agent for QA to interactively search the tree to find relevant information.
Our system, \ours (Hierarchical Episodic Memory Verbalization, \cref{fig:title}), processes the continuous stream of experiences and inserts it into a hierarchical representation of the robot's history of episodic events.
Different levels of this hierarchy represent different abstraction levels, with the lowest level being raw observations and proprioception and higher levels being represented as natural language concepts.
An LLM is prompted for segmentation and summarization in order to recursively create higher-level abstractions.
To process queries to the EM, we repurpose the interactive prompting scheme described in our previous work~\cite{barmann_incremental_2024}.
An LLM is provided with the user's query and functions to access the history tree, and the LLM itself decides which functions to use in order to fulfill the query, \ie to answer the question or provide a summary.
Since the history tree will grow large over time, we apply an interactive semantic search, inspired by work on robot navigation \cite{rana_sayplan_2023,xie_embodied-rag_2025}.
Specifically, we contract the tree, \ie the LLM initially only sees the top-level node, and then interactively explores the graph to retrieve the information relevant to the query.

To evaluate our system, we use simulated household episodes from TEACh~\cite{padmakumar_teach_2022}, real-world egocentric human recordings from Ego4D~\cite{grauman_ego4d_2022}, and real-world robot episodes from \armarVII, the newest member of the \armar humanoid robot family developed at KIT \cite{asfour_armar_2017}.
Our experiments show that \ours efficiently scales to extremely long histories of multiple simulated months or real-world human egocentric videos of over six hours, outperforming several baselines and ablations.
Real-world robot demonstrations showcase the wide applicability of our system.
We provide our code, evaluation data, and demonstration videos at \href{https://hierarchical-emv.github.io}{hierarchical-emv.github.io}.

\section{Related Work}

\paragraphHeading{Episodic Memory for Robots}
The concept of EM stems from human cognition \cite{tulving_episodic_1972} and is useful for various technologies including smart wearables \cite{grauman_ego4d_2022,barmann_where_2022}, smart rooms \cite{waibel_chil_2005}, and especially robotics.
For instance, robotic EM can be represented using latent vectors created by deep neural models \cite{barmann_deep_2021,rothfuss_deep_2018,dechantsearch_2024}, or by explicitly storing relevant information in a memory system \cite{Peller-Konrad2023MemorySystemRobot,beetz_knowrob_2018,plewnia_forgetting_2024}.
Another approach is to represent the history of episodic events as text produced by pretrained models \cite{zeng_socratic_2023,liu_reflect_2023}.
In this paper, we also represent the history tree in form of text, following REFLECT~\cite{liu_reflect_2023} for the broad structure of the hierarchy's lower levels.
However, we extend this by adding hierarchical summarization.
Furthermore, our multimodal episodic history tree can be dynamically explored by an LLM to gather information from all levels, including the raw observations.

\paragraphHeading{Robot Experience Verbalization}
The first work to introduce the term of \enquote{verbalizing} robot experiences was \cite{rosenthal_verbalization_2016}.
With a rule-based system, they converted a navigation route taken by a mobile service robot to natural language.
\cite{zhu_autonomous_2017} adapted this framework to verbalization of manipulation activities performed by a humanoid household robot.
Similarly, \cite{yuguchi_butsukusa_2022} use templates to convert their robot's observations and actions to natural language.
More recent works phrase EMV in a more interactive setting, defined as summarization and QA on robot experiences \cite{dechant_toward_2021}.
Both \cite{barmann_deep_2021,dechant_learning_2023} propose end-to-end trained networks receiving multimodal experiences and a question to produce an answer, using training data from simulated household tasks.
While \cite{dechant_learning_2023} work on visual data only, \cite{barmann_deep_2021} additionally use symbolic and subsymbolic information from the robot's task execution and perception components.
In contrast to these systems, \ours uses pretrained foundation models and does not require additional training data, thus increasing its versatility and easing deployment to the real world.
Most similar to our work, ReMEmbR \cite{anwar_remembr_2025} propose to use a robot's experience for QA and navigation.
Using an LLM as an agent to decide what data to retrieve from a vector store of memories, they can answer queries based on time, place, and content.
The \enquote{search} function available to \ours's LLM agent works similar to this but acts on a specific tree node instead of the overall history.
Further, ReMEmbR's memory is not hierarchical, thus limiting both temporal and content-based retrieval to short moments in time, struggling to handle queries that require larger context (such as \enquote{summarize your day}).
Focusing on failure identification and recovery, \cite{wang_i_2024} produce robot narrations from multimodal robotic data streams.
This can be seen as orthogonal to our work as these narrations could enrich the lower levels of our hierarchy.
Similar to our setting, QA from streaming data \cite{han_episodic_2019,araujo_memory_2023} tackles the problem of answering questions based on a long stream of data where the question is not known in advance and the raw data cannot be stored.
However, we apply this to robotics, and approach it with an interpretable, modular system instead of end-to-end trained memory models.
Furthermore, some works in natural language processing propose hierarchical variants of retrieval-augmented generation to tackle multi-document QA tasks \cite{wang_archrag_2025,edge_local_2025,chen_hiqa_2024}, which is similar to our work in that a graph-like structure is accessed, but differs as we use an LLM as an agent and apply it on inherently temporal, multimodal robotics data.

\paragraphHeading{Video Understanding} Video Understanding, especially Video Question Answering (VideoQA), is related to EMV as it also involves QA on a data stream, which, however, is only a video instead of a multimodal robotic experience stream.
VideoQA is an active research area \cite{zhong_video_2022} where current major challenges include long-form videos beyond clips of a few seconds as well as egocentric video understanding.
Ego4D \cite{grauman_ego4d_2022} is a large collection of unconstrained egocentric videos showing daily activities of human camera wearers.
Ego4D GoalStep \cite{song_ego4d_2023} and HCap \cite{islam_video_2024} provide hierarchical annotations for subsets of Ego4D, facilitating reasoning on different abstraction levels.
Recent long-form egocentric VideoQA benchmarks include \textsc{QaEgo4D} \cite{barmann_where_2022} and EgoSchema \cite{mangalam_egoschema_2023}.

Recent methods for VideoQA can be grouped into (i) end-to-end approaches \cite{balazevic_memory_2024,song_moviechat_2024,tan_koala_2024,islam_video_2024,di_grounded_2024} that typically connect pretrained frozen visual encoders with LLMs by some trained adapter, and (ii) training-free \enquote{socratic} \cite{zeng_socratic_2023} approaches \cite{fan_videoagent_2024,wang_videotree_2024,wang_videoagent_2024,min_morevqa_2024,wang_lifelongmemory_2023,choudhury_zero-shot_2023,zhang_simple_2024} that invoke various off-the-shelf models to convert the video into text to be processed by a few-/zero-shot prompted LLM.
For instance, \cite{wang_lifelongmemory_2023,zhang_simple_2024} use video captioning to produce a transcript of the video and then apply an LLM for QA based on this transcript.
VideoTree \cite{wang_videotree_2024} adaptively selects the frames to caption using a top-down query-relevance-based tree expansion instead of uniform sampling.
Both \cite{huang_building_2025,goletto_amego_2025} build structured representations from egocentric video using various models, \eg for object and hand tracking, as well as activity and location labeling.
\cite{choudhury_zero-shot_2023} generate executable Python code from a question, invoking different APIs to query visual and language foundation models.
MoReVQA \cite{min_morevqa_2024} decomposes this into multiple stages, making the LLM's job easier at each stage by focusing on either event parsing, grounding, or reasoning, instead of all at once.
In contrast to these predefined prompting schemes, both \cite{fan_videoagent_2024,wang_videoagent_2024} use an LLM as an agent to analyze the video content in an interactive loop.
While \cite{wang_videoagent_2024} iteratively ask the LLM whether to gather more detailed information (by captioning more intermediate frames) or produce the final answer, \cite{fan_videoagent_2024} provide the LLM with API functions invoking tools to search in a database of tracked objects or a memory of frame captions.

Our method similarly treats the LLM as an agent, thus not relying on any predefined information flow.
However, we use the full flexibility of code \cite{wang_executable_2024} instead of single API calls like in \cite{fan_videoagent_2024,wang_videoagent_2024}.
Compared to VideoTree \cite{wang_videotree_2024}, our history tree is constructed independently of the user's query, since future questions cannot be known in advance in realistic settings, and storing lifelong \enquote{raw} video experiences is prohibitive \cite{barmann_where_2022}.
In contrast to all of the above works, we consider real-world dates and times an integral part of the process.
While the recent work TimeChat \cite{ren_timechat_2024} is also time-sensitive, they refer to video timestamps (\enquote{second 42 to 57}) instead of real-world date-times (\enquote{yesterday in the afternoon}).
Furthermore, and most crucially, we deal with long sequences of multimodal \emph{robotic} experiences, with the longest experiment having over six hours of video or nearly two months on a simulated timeline.

\section{Method}

Our goal is to enable an artificial agent to verbalize and answer questions about its past.
Given the continuous, multimodal stream of experiences of a robot agent, we build up a hierarchical and interpretable representation of EM (\cref{sec:method:em_construction}).
When a user later asks a question, an LLM interactively explores the history tree to gather relevant information, detailed in \cref{sec:method:em_access}.

\subsection{Episodic Memory Construction}
\label{sec:method:em_construction}
\newcommand{\treelevel}[2]{\textbf{L#1 -- #2:}}

\begin{figure}
    \centering
    \includegraphics[width=1\linewidth]{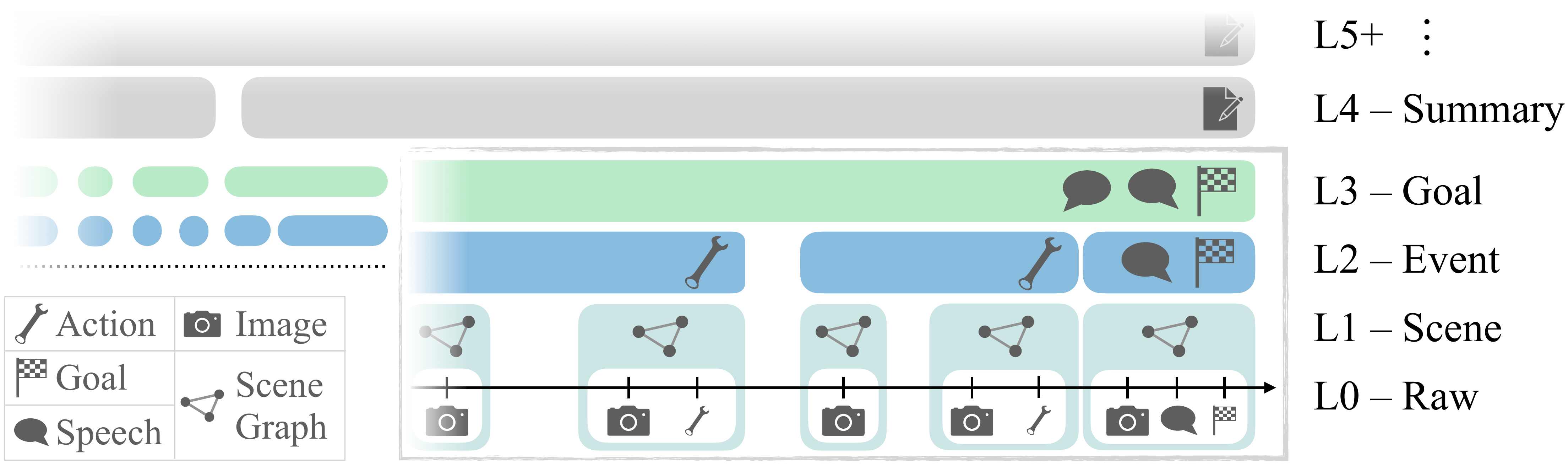}
    \caption{From the continuous, multimodal stream of robotic experiences, we construct a \emph{history tree}, a hierarchical representation of the EM.}
    \label{fig:em-tree}
\end{figure}

From a stream of multimodal robot experiences, we derive a hierarchical representation of the robot's EM, a \emph{history tree}, as shown in \cref{fig:em-tree}, with the lower levels broadly following~\cite{liu_reflect_2023}.
Specifically, the tree's levels are:

\treelevel{0}{Raw Experiences} 
Leaf nodes collect the raw information available at a specific timestep during the robot's task execution.
This includes all modalities that can be perceived by the robot: RGB and depth camera images and recorded audio, as well as information deduced from this data, \ie recognized objects, their positions, and a text transcription of the audio, if there is user speech.
Furthermore, we include everything the agent knows about its state: robot proprioception (joint configuration, mobile platform position), symbolic information about the current action and goal, and text to be spoken by the robot's text-to-speech component.

\treelevel{1}{Scene Graphs}
The first level of non-leaf nodes in the history tree has a one-to-one mapping to the L0 leafs.
On this level, we derive a scene graph from the given observations, consisting of the detected objects as nodes and their spatial relations (\eg on top, inside) as edges.
The exact method for constructing the scene graph varies in our experiments.
For the pure vision-based approach, objects are detected  using pretrained models and heuristics are applied to infer semantically meaningful relations \cite{liu_reflect_2023}, whereas
in our real-robot experiments, we use the existing components in our robot's software framework \armarx~\cite{vahrenkamp_robot_2015} that already provide semantic scene information.

\treelevel{2}{Events}
Next, we group and summarize the nodes from the previous level based on changes in the scene graph, the currently executed action or goal, as well as when there is a new speech recognition.
We also create a template-based natural-language summary, including the latest scene graph, the current action, and recognized speech command.
In our real-robot experiments, we use an LLM to filter and summarize the raw action parameters, which would be excessively detailed otherwise.

\treelevel{3}{Goals}
Based on the current goal from the L0 node, we group event nodes and again create a rule-based natural-language summary containing the current goal and the verbalization of the latest event.
Note that we allow goal nodes to have children of mixed types: either events or other goal nodes.
This allows representing subgoals of complex tasks and is used in our real-robot experiments.

\treelevel{4+}{Higher-Level Summaries}\label{sec:methods:higher_lvl_summary}
Summaries are generated dynamically by recursively asking an LLM to summarize the previous level's nodes.
Specifically, given the set of nodes $S_\ell$ at level $\ell\geq3$, we list them in order and prompt an LLM to identify consecutive ranges of items that belong together considering their times and content, and provide a summary for each range.
The output is parsed to group the child nodes and create the summary nodes $S_{\ell+1}$ for the next level.
We apply this strategy recursively, until $|S_\ell|=1$ or there is no further reduction, \ie $|S_{\ell+1}| = |S_\ell|$.
In the latter, the LLM is explicitly prompted to provide a single concise summary of all items to force obtaining one root node.
To make this procedure robust to LLM errors, \eg invalid output syntax, missing an item or including one in multiple ranges, we re-prompt on failure, listing detailed error messages and asking for improvement.
In case the LLM fails to provide a valid grouping after three trials, we fall back to the single summary prompting.

\subsection{Episodic Memory Access}
\label{sec:method:em_access}

\begin{figure*}
    \centering
    \includegraphics[width=1\linewidth]{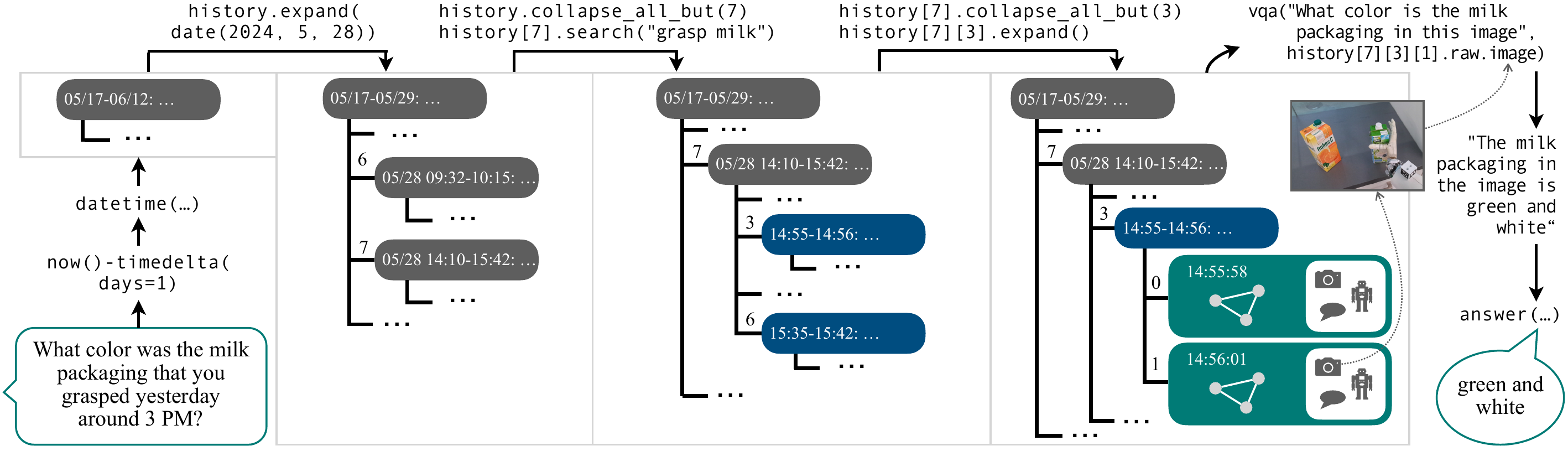}
    \caption{To answer a user's question, \ours prompts an LLM to interactively explore the history tree containing the agent's experiences. The LLM can further invoke tools (index search, VLM) or perform other calculations to gather relevant information, eventually invoking the \texttt{answer} function. This figure shows an example from our real-world evaluation on the humanoid robot \armarVII, modified for illustrative purposes.}
    \label{fig:tree-search}
\end{figure*}

Given a user's query and the history tree built from all experiences so far, we use an LLM as an agent \cite{wang_survey_2024} to explore the tree, search relevant information, and eventually answer the question.
For this, we define an API to interact with the history tree.
We initially define each node of the tree to be in a collapsed state, \ie its textual representation will only contain the node's time range and natural-language summary, but not list the child nodes.
The LLM can then interactively expand and collapse nodes, according to what seems relevant given the user's query.
Furthermore, we provide different tools to the LLM, \eg to invoke a Vision-Language-Model (VLM) to perform visual QA on the images associated with leaf nodes.
Moreover, there is a function to perform tree search based on semantic similarity, selectively expanding the children of the searched node in the tree that match the search query.

\Cref{fig:tree-search} illustrates typical steps the LLM performs to answer a user's question.
Given the initially collapsed tree, the LLM first expands the root node's children based on the requested date.
It then selectively explores the respective child nodes that seem relevant to the question using the search function.
Note that the LLM is prompted to collapse irrelevant nodes again in order to limit the number of tokens used and speed up further requests.
In the given example, when reaching a leaf node, the answer to the question is not evident from any of the natural-language summaries on each level so the LLM decides to invoke a VLM to gather more information.
Finally, it invokes the \texttt{answer} function to answer the user's question.

Our implementation of the LLM agent uses a prompting style inspired by the simulated Python console of \cite{barmann_incremental_2024}.
The LLM can issue any command -- including compound statements such as loops -- using the provided API.
After the execution of the respective code, the LLM can \enquote{see} the output of its command(s), or any execution error.
This process is repeated, and the prompt to the LLM always contains the (growing) execution history.
Our API returns detailed error messages and usage hints in case the LLM uses it in a wrong way, and we also detect repetitive generations, prompting the LLM to try to solve the problem differently.
In case an answer is not obtained after a set amount of maximum iterations, we use a special prompt to force an immediate answer given the current state of the history tree.
Zero-shot experiments prompt the LLM with only a static prefix to explain the task and the available API, while few-shot prompting adds top-$k$ examples selected based on semantic similarity to the current user query.
The final part of the prompt is always a string representation of the history tree's current state.

\section{Evaluation}

\subsection{Simulated Household Episodes}
\label{sec:eval:teach}
Following our previous work~\cite{dechant_learning_2023,barmann_deep_2021}, we use simulated household episodes and automatically annotate them with QA pairs based on the ground-truth (GT) simulation state.
Specifically, we use the TEACh dataset~\cite{padmakumar_teach_2022}, featuring episodes of two real humans, one commander, and one follower, interacting with the AI2THOR environment~\cite{kolve_ai2thor_2017}.
We adapt this data by rephrasing the commander to be a human user, and the follower to be a robot interacting with the environment (and the user).
Thus, each TEACh episode describes a robot experience, comprising egocentric images, robot states and actions, and dialog with the user.

\paragraphHeading{Data}
Episodes in TEACh are on average $6.2 \pm 5.3$ min long.
Since we are interested in very long histories of robot experience, we randomly combine them to form histories of up to 100 episodes.
We also randomize dates and times for each episode, ensuring realistic sequences by picking one to five episodes per day, avoiding nighttimes, and occasionally skipping some days; the longest histories thus span nearly two months.
Based on these histories, we generate QA pairs by adjusting the generation grammar from \cite{dechant_learning_2023}.
Specifically, we generate ten types of questions. These ask for: a list of high-level summaries of episodes (task descriptions); a detailed description of one particular episode in a history; a summary of an episode that happened either before or after a particular episode; a list of episodes in which a particular object was seen or action performed; a summary of an episode that occurred at a given time or a specified number of days ago; a list of times or number of days ago at which a given task was performed.
Questions can refer to specific episodes either by date and time, or by the task performed (if it is unique).
From the TEACh \enquote{valid unseen} set, we generate test sets with 10 histories per sequence length (combining~$|h|=5,\, 15,\, 25,\, 50$, and $100$ episodes).
Each history $h$ is annotated with ten QA pairs, making up 100 samples per history length $|h|$.

\newcommand{\teachTabHeading}[1]{\multicolumn{5}{c|}{#1}}
\newcommand{\teachTabHeadingN}[1]{\multicolumn{3}{c|}{#1}}
\newcommand{\teachmetrics}[0]{B & R & $S_c$ & $S_p$ & T }
\newcommand{\teachmetricsN}[0]{ $S_c$ & $S_p$ & T }

\begin{figure*}
\begin{minipage}{0.6\textwidth}
        \setlength\tabcolsep{2px}
    \renewcommand{\arraystretch}{1}
    \scriptsize
        \captionof{table}{Results on Simulated Household Episodes from TEACh \cite{padmakumar_teach_2022}}
        \vspace{-0.3cm}
        \begin{tabular}{rrr|rrr|rrr|rrr|rrr|rrr}
            \multicolumn{3}{c|}{$\to |h|$}  &  \teachTabHeadingN{5} & \teachTabHeadingN{15} & \teachTabHeadingN{25} & \teachTabHeadingN{50} & \multicolumn{3}{c}{100}\\
            method & ICL & Hier.   & \teachmetricsN & \teachmetricsN & \teachmetricsN & \teachmetricsN & \teachmetricsN \\ \hline
            \multicolumn{18}{l}{vision-only} \\\hline
            img 1-pass  & FS & flat &  24 &  49 & 164 & 13 &  43 & 334 &  17 &  41 & 406 &  11 &  35 & 512 & 10 &  30 & 1256 \\
            text 1-pass & 0  & L3   &   5 &  18 & 65.4 &  8 &  15 & 336 & 5 &  15 & 270 &  5 &  25 & 789 & 6 &  14 & 1234\\
            \ours       & 1  & full &   9 &  40 & 10.6 &  9 &  37 & 10.2 &  10 &  42 & 10.2 &  15 &  36 & 11.6 &  15 &  35 & 11.9  \\\hline
            \multicolumn{18}{l}{vision + speech} \\\hline
            img 1-pass  & FS & flat &  57 &  83 & 165 & 30 &  73 & 336 &  39 &  69 & 418 & 25 &  60 & 861 & 17 &  54 & 1591 \\
            text 1-pass & 0  & L3   &  21 &  43 & 68.9 & 19 &  43 & 217 & 22 &  50 & 287 & 15 &  41 & 740 &  7 &  28 & 1304 \\
            \ours       & 1  & full &  28 &  64 & 9.2 &  33 &  63 & 9.9 &  31 &  63 & 9.7 &  27 &  57 & 10.9 & 19 &  51 & 13.3 \\\hline
            \multicolumn{18}{l}{full multimodal (objects + speech + actions)} \\\hline
            ReMEmbR     &    &      &  10 &  30 & 12.2 &  6 &  21 & 12.5 &  11 &  28 & 12.5 &   7 &  22 & 12.6 &  5 &  18 & 12.7  \\
            text 1-pass & 0  & L3   &  32 &  65 & 120  &  37 &  63 & 372  &  35 &  69 & 422  &  24 &  56 & 1055 &  \multicolumn{3}{c}{\textcolor{gray}{OOC}} \\
            \ours       & 0  & L3   &  28 &  61 & 19.7 &  20 &  62 & 44.2 &  29 &  65 & 19.8 &  16 &  64 &  116 & 21 &  52 & 249 \\
            \ours       & 0  & full &  51 &  71 & 3.1  &  45 &  76 & 4.1  &  46 &  73 & 15.0 &  32 &  58 &  4.9 & 25 &  60 & 5.6 \\
            \ours       & 1  & full &  48 &  74 & 8.5  &  48 &  72 & 9.6  &  55 &  74 & 23.7 &  37 &  68 & 10.2 & 34 &  62 & 10.4\\
            \hline
             \multicolumn{18}{c}{}\\[-1em]
             \multicolumn{18}{c}{$S_c, S_p$: semantically categorized (partially) correct in \%, T: number of 1K prompt token.}\\
             \multicolumn{18}{c}{OOC: out of context, ICL: in-context learning, FS: few-shot}
    \end{tabular}
    \label{tab:results:teach}
\end{minipage}
\begin{minipage}{0.39\textwidth}
    \input{figures/tikz/tokens_vs_performance_per_ep_len.tikz}
    \captionof{figure}{Token costs vs. performance for different history lengths (TEACh full multimodal). Solid lines are $S_p$, dashed lines $S_c$. \ours retains better performance at lower costs.}
    \label{fig:results:tokens_vs_perf}
\end{minipage}
\end{figure*}

\paragraphHeading{Evaluation Metrics}
\label{sec:eval:teach:metrics}
Evaluation of free-form EMV answers is hard since there can be many ways to formulate the correct answer, questions can be underspecified, and verifying abstract statements by grounding them in the history tree is a research question in itself.
Following \cite{barmann_deep_2021}, we define a semantic categorization of a model's output $o$ given the GT $g$ and question $q$:
\emph{correct} when $o$ is semantically equivalent to $g$;
\emph{correctly summarized} if $o$ is a correctly summarized version of $g$, still containing all relevant facts (in context of $q$);
\emph{correct TMI} (too much information) if $o$ is correct but overly specific;
\emph{partially correct TMI} if parts of $o$ are correct, but there are TMI parts and these are wrong;
\emph{partially correct missing} if parts of $o$ are correct, but relevant facts from $g$ are missing;
\emph{wrong} when $o$ could be an answer to $q$ but is none of the above;
and \emph{no answer} if $o$ is empty, completely irrelevant to $q$, or the model threw an error.
Counting \textit{correct} (incl. correct summarized, correct TMI) samples $C$ and \textit{at least partially correct} samples $P$ ($C\subseteq P$), we report their percentages $S_c$ and $S_p$, respectively.
Since categorizing each evaluated sample by hand is prohibitively expensive, we prompt GPT-4o~\cite{openai_gpt4_2023} to perform this evaluation.
For this, we started by annotating 60 samples by hand, and use these as a database to retrieve few-shot samples based on maximal marginal relevance~\cite{ye_complementary_2023}.
We further tuned the prompts on a validation set of 100 hand-categorized model outputs (generated from TEACh train).

We evaluate the agreement of the LLM's predicted categories with manually annotated ones on 200 model results from our test data, resulting in an aggregated category accuracy of $88\%$, and per-class f-scores of $F_1(\text{correct})=0.89, F_1(\text{partially\_correct})=0.84, F_1(\text{wrong})=0.91$.
The LLM categorizes correct and wrong samples very well and has the most difficulties on the \emph{partially correct} labels.
However, these categories are also defined imprecisely, and the inter-annotator agreement~\cite{cohen_coefficient_1960} between the first two authors has only a value of Cohen's $\kappa=0.66$ ($n=110$), vs. $\kappa=0.91$ ($n=68$) when only considering correct/wrong.
Thus, while not perfect, we use the LLM to automatically obtain reasonable score estimates.
We do not report surface-level automated metrics such as BLEU \cite{papineni_bleu_2002} or ROUGE \cite{lin_rouge_2004} as these metrics struggle to capture the variety of correct answers in the EMV task, \eg for a \enquote{when}-question, both of the following answers are correct, but have no word overlap at all: \enquote{at 4 PM} vs. \enquote{in the afternoon}.

\paragraphHeading{Settings and Baselines}
We evaluate under three settings:
First, \underline{\smash{vision-only}} can act solely on the visual data stream.
As a baseline, we prompt Gemini 1.5 Pro \cite{gemini_team_gemini_2024} in one pass with the sequence of images along with timestamps and the question (\textit{img 1-pass}).
We choose Gemini because we require extremely long requests; however, despite its 2M token limit, we need to sample every 2nd frame so that longer histories fit into the context window.
We few-shot-prompt with one static example history of five episodes including ten QA samples for this history.
For a fair comparison, we use the same LLM for ReMEmbR and \ours (although we can achieve better results with GPT-4o, as shown in \cref{sec:eval:robot}).
\ours does not take raw images, but constructs history trees by inferring objects using YOLO-World \cite{cheng_yolo-world_2024} and actions using a LongT5 transformer model~\cite{guo_longt5_2022} fine-tuned on TEACh train. 
This vision model limits our system with an action classification accuracy of $62\%$ (which can be partially circumvented by the summarization LLM's commonsense knowledge).
Few-shot prompting Gemini to infer the sequence of actions for building the tree had a worse accuracy of around $27\%$.
We also compare with one-pass zero-shot prompting the LLM with this tree limited to L3 node (\ie no hierarchical summarization) in text form (\textit{text 1-pass}).
Second, \underline{\smash{vision + speech}} enriches the visual information with the dialog data from TEACh episodes, representing natural language commands given to the robot.
This is simply added to the prompt for \textit{img 1-pass}, and inserted into the history tree for \ours and \textit{text 1-pass}.
Finally, \underline{full multimodal} uses the recorded (GT) actions and goals from the TEACh episodes, as this information is typically available when a robot executes some actions.
This setting also uses GT object information to compare system performance assuming perfect vision components.
We compare \ours with one-shot and zero-shot prompting of the LLM agent.
For preparing the few-shot samples, we use histories built from episodes in TEACh train, and record traces of manually using the Python console interface and the defined API to interact with the history tree until the given GT answer becomes evident.
While we collect two to three samples per question type this way, making up 21 samples in total, we select only the top-1 sample when prompting the LLM, based on semantic similarity of the user's questions.
Semantic similarity is determined after asking \texttt{gpt-4o-mini} to cross out the task-specific words from the question so that an example from the same question type is retrieved (instead of an irrelevant example just mentioning the same objects or activities).
We further ablate the hierarchical summarization: \textit{\ours 0-shot L3} is our method without LLM-generated summaries (L4+), still using the interactive agent to explore an initially collapsed list of L3 nodes.
Last, \textit{text 1-pass 0-shot L3} is a baseline presenting the fully expanded tree (L3 and lower) to the LLM along with the question in a single prompt, thus not using the LLM as an agent.
We also compare to ReMEmbR \cite{anwar_remembr_2025}, with their method/prompting minimally adjusted to our setting.

\paragraphHeading{Results}
Results of our TEACh experiments can be found in \cref{tab:results:teach}.
First, we can observe that every method's performance decreases with increasing $|h|$.

Focusing on the vision-only and vision + speech results, the Gemini 1-pass baseline outperforms \ours for shorter histories.
This is reasonable, as the baseline can directly access the full stream of visual information, whereas our hierarchical system suffers from error propagation and is limited by pretrained vision components.
In particular, the history tree could contain incomplete or wrong information or our method could fail by expanding the wrong nodes of the tree, which cannot happen to the 1-pass baseline.
However, token costs scale linearly with history length for 1-pass, while it stays approximately constant for \ours. 
The performance also drops faster for 1-pass, with \ours reaching comparable or better semantic scores for $|h| \geq 25$.
In contrast, when circumventing the limitations of perception components by using GT object detection and action information (full multimodal setting), \ours outperforms the 1-pass system with a token budget two orders of magnitudes smaller (see \cref{fig:results:tokens_vs_perf}).
Further, 1-shot prompting helps for longer histories, but 0-shot works with half the token costs and comparable performance for short histories.
Ablating the hierarchical higher-level summaries notably increases token cost and leads to worse performance, especially for longer histories (when the LLM sometimes expands all nodes).
While ReMEmbR keeps token cost low due to its use of the LLM as an agent similar to \ours, its retrieval-augmented generation approach and lack of summarization capability has difficulties with our QA data, as it typically retrieves single actions (\enquote{I picked up a mug}) instead of higher-level task information (\enquote{I prepared breakfast}).

\subsection{Egocentric Human Videos}
\label{sec:eval:ego4d}
In addition to verbalizing robot experience, EMV can be applied to human egocentric recordings, \eg in the context of smart wearables.
Here, the system does not summarize and answer questions about its own actions, but the actions of its user.

\paragraphHeading{Data}
To evaluate our system under this setting, we use Ego4D~\cite{grauman_ego4d_2022}.
Randomly concatenating episodes (as done above for TEACh) generates histories that are not cohesive, thus restricting automatic summarization to bare enumeration instead of abstraction of related events.
Therefore, we perform a small-scale evaluation on very long recordings from Ego4D.
Specifically, we manually select two very long Ego4D videos (6:43h and 4:28h) showing diverse and interesting actions in a tourist scenario.
Additionally, we construct one history by concatenating shorter episodes from similar scenarios, selected to ensure some level of cohesiveness and plausibility (in contrast to random sequences).
We manually write 40 challenging QA samples.

\paragraphHeading{Method}
To construct history trees from Ego4D videos, we apply VideoReCap \cite{islam_video_2024} which produces low-level narrations at 1 fps and mid-level summaries for each minute.
We map these to action (L2) and goal (L3) nodes of our hierarchy, respectively, converting texts to first-person perspective using \texttt{meta-llama3-8b}~\cite{dubey_llama_2024}.
For constructing higher-level (L4+) summaries, we generated few-shot samples for the group-and-summarize LLM (see \cref{sec:methods:higher_lvl_summary}) using Ego4D-HCap \cite{islam_video_2024}. 
To populate the L1 scene graph with objects, we apply YOLO-World \cite{cheng_yolo-world_2024}, an open-vocabulary object detection approach, which we prompt with classes obtained through a Socratic Models~\cite{zeng_socratic_2023} approach: 
First, we select the top-100 classes according to cosine similarity of the mean CLIP \cite{radford_learning_2021} image embedding within one L3 node and the CLIP text embeddings of all LVIS \cite{gupta_lvis_2019} labels.
Further, we prompt \texttt{meta-llama3-8b} to propose $\approx 20$ objects that might occur given the current L3 goal annotation produced by VideoReCap.
We then apply YOLO-World with the combined set of classes on each image within this L3 node and store the detected objects in the respective L1 nodes.
The EMV agent for QA is instantiated zero-shot.

\paragraphHeading{Results} The results of applying our method and manually categorizing the results (following \cref{sec:eval:teach:metrics}) can be seen on the left of \cref{tab:results:real-world}.
Due to the very challenging nature of our QA samples and the limitations of the used vision components, the overall performance is low.
Low performance of the 1-pass baseline can partially be explained by it seeing only a flat version of the L2 events without access to the images.
However, it also fails on most of the samples that could be answered from the text history. 
This may be explained by the noisy text history inferred from vision, which \ours can handle better because of hierarchical summarization and selective expansion of nodes, whereas the 1-pass baseline observes all the noise at once.
We did not directly apply the 1-pass baseline on the images (which would be possible only with aggressive sub-sampling) for cost reasons.
\ours L3 performs slightly worse than \ours, with double the costs.

\begin{figure*}
    \centering
    \includegraphics[width=1\linewidth]{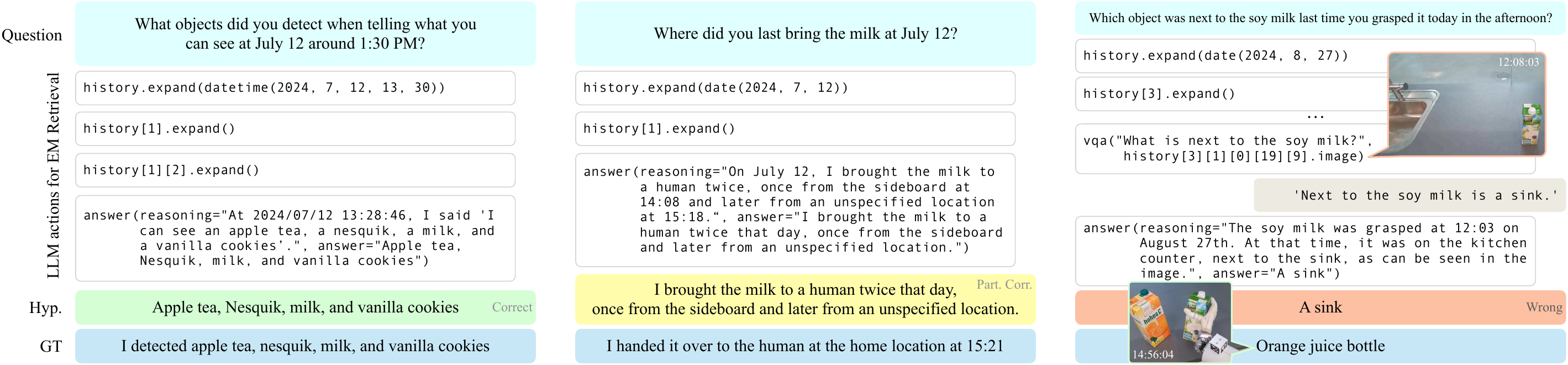}
    \caption{Example traces from our \armarVII evaluation. \emph{Left}: success case. \emph{Middle}: partially correct, LLM misinterpreting the question. \emph{Right} (shortened): wrong answer, expanded the wrong node (not the \emph{last} time grasping the milk). More examples at \href{https://hierarchical-emv.github.io}{hierarchical-emv.github.io}
    }
    \label{fig:real-world-samples}
\end{figure*}

\subsection{Real-World Robot Recordings}
\label{sec:eval:robot}

Finally, we apply our method on our real-world humanoid robot \armarVII.
To obtain an EM, we record multiple robot sessions of typical household tasks, spanning a total duration of 3.3 hours of robot actions over the scope of two months.
We record all entries made to the memory system introduced in \cite{Peller-Konrad2023MemorySystemRobot}, in particular: vision (RGB and depth images), robot state (proprioception), skill events (executed actions and goals), speech (speech-to-text output and text-to-speech input), symbolic scene (objects and their relations).
From such recordings, we build up a history tree by populating L0 with images, speech, and proprioception, L1 scene graphs with the symbolic scene information, L2 and L3 with robot action events (where L2 contains low-level actions and L3 contains actions that themselves invoke other actions).
Note that L3 nodes can be nested in this case (goals and their subgoals).
Higher levels (L4+) are constructed dynamically by an LLM as described in \cref{sec:methods:higher_lvl_summary}, with two manually created few-shot samples.

\begin{table}[t]
    \setlength\tabcolsep{3.5px}
    \renewcommand{\arraystretch}{1}
    \scriptsize
    \centering
    \caption{Experiments on Human \& Robotic Real-World Data (0-shot)}
    \begin{tabular}{rrr|rrr||rrr}
               &           &        & \multicolumn{3}{c||}{Ego4D} & \multicolumn{3}{c}{\armarVII}   \\ \hline
        Method & Hierarchy & LLM    & $S_c$ & $S_p$ & T& $S_c$ & $S_p$ & T   \\ \hline
        1-pass & flat      & Gemini &   5 &  13 &  438 &  43 &  53 & 227 \\
        \ours  & L3        & Gemini &  18 &  38 & 33.1 &  40 &  70 & 29.2 \\
        \ours  & full      & Gemini &  30 &  40 & 14.4 &  33 &  73 & 10.5 \\
        \hline
        \ours  & L3        & GPT-4o &  25 &  43 & 57.6 &  30 &  47 & 68.4 \\
        \ours  & full      & GPT-4o &  28 &  53 & 21.9 &  43 &  70 & 12.8 \\
    \end{tabular}
    \label{tab:results:real-world}
\end{table}

Subsequently, we annotate the recordings with 30 QA-pairs, apply our method, and again manually categorize the results.
Results can be seen on the right part of \cref{tab:results:real-world}, with examples in \cref{fig:real-world-samples}.
In general, our task is very challenging, and the 1-pass Gemini baseline which has direct access to the complete stream of episodic data (without images) scores only 43\%/53\% of correct/partially correct samples.
Compared to the Ego4D experiment, the quality of the text history is better, as most content (esp. current action, goal) is not inferred from vision.
Our interactive hierarchical system achieves better $S_p$ and slightly worse $S_c$, with 1/21 of the token costs.
While the experiments with Gemini and limited tree depth (L3) perform slightly better, the best results are achieved by the full system with GPT-4o. Here, L3 has lower performance with more than five times the token cost.
See the supplementary video for a demonstration of our system in action, enabling \armarVII to answer questions about its past interactively.
We also applied \ours (with no further dataset-specific prompt tuning) on NaVQA \cite{anwar_remembr_2025}, but obtained an accuracy of only 32\% on the descriptive questions, which we attribute to their dataset being focused on very specific details, whereas our method strives at tasks that require broader context and summarization capabilities.

\subsection{Failure analysis}

To better understand the potential for future improvement of our system, we analyze its failure modes.
We broadly categorize failures by three sources: EM construction, EM retrieval, and LLM QA reasoning.
For the simulated TEACh evaluation, full multimodal setting, both the generated QA and the history tree are derived from the simulation state, therefore the requested information is guaranteed to be contained in the tree, eliminating the EM construction failure reason.
During QA generation, we also annotate the target episode this question asks for, and then evaluate whether our model correctly expands the history tree node representing this episode.
Specifically, we calculate retrieval precision $p$ and recall $r$ as follows:
Given the final state of the tree when the LLM decides to answer the question, we extract the set of visible (expanded) nodes $V$.
From these, we pick the lowest-level visible nodes: $L = \{v \in V \ |\ \text{children}(v) \cap V = \varnothing \}$.
Every node $v$ has an associated time span $t_v = (s_v, e_v)$ with start and end datetime $s_v, e_v$.
Given a reference time span $t_g = (s_g, e_g)$ that the question refers to (one question may refer to multiple time spans, \eg \enquote{what did you do before...}), we calculate the length of the intersection $i_{g,v} = |t_g \cap t_v|$.
This intersection is normalized relative to the duration of the node and the baseline intersection $b$ of the root node $v_r$ with the reference, defined as $b = \frac{|t_g|}{|t_{v_r}|}$.
The normalized intersection is given by $n_{g,v} = \max(0, \frac{i_{g,v}}{|t_v|} - b) \cdot \frac{1}{1-b}$.
Recall is defined as the average of the maximum normalized intersection rate per reference time span: $r = \text{avg}_{t_g \in G}(\ \max_{v\in V} n_{g,v}\ )$.
Precision is calculated by taking the maximum normalized intersection rate for each leaf $l \in L$ over all reference spans, and then averaging across all leaves: $p = \text{avg}_{l \in L}(\ \max_{t_g \in G} n_{g,l} \ )$.
Eventually, $F_1 = \frac{p \cdot r}{p + r}$.

\Cref{fig:results:retrieval_f1_1shot} shows the retrieval $F_1$ of \ours for different history lengths.
Evidently, the retrieval process is a major source of problems for the EMV task:
When taking into account only samples that were answered (partially) correct, the retrieval performance is notably better than for the samples that were answered wrong, underlining that correct retrieval is essential for overall QA performance.
The non-zero retrieval performance for wrong samples stems from partial retrieval (not expanded deep enough) as well as other failures due to QA reasoning (despite correct retrieval).

\begin{figure}
	\centering
	\input{figures/tikz/teach-retrieval-f1-1shot.tikz}
	\caption{Retrieval $F_1$ score by history length $|h|$ (TEACh \ours 1-shot)}
	\label{fig:results:retrieval_f1_1shot}
\end{figure}

For our real-world (Ego4D and \armarVII) experiments, we manually inspect the execution traces of \ours (Gemini) for each (partially) wrong sample and attribute it with one or multiple of the above failure reasons.
As for the simulated evaluation, the majority of failures is related to retrieval (18 of 28 failure samples for Ego4D, 19/20 for \armarVII).
Especially for the real-robot experiments, the tree is very deeply nested, and retrieval therefore challenging.
On the other hand, tree building issues are more prevalent for Ego4D (13/28), as the system is relying purely on visual information there.
However, issues can also occur on \armarVII (7/20), for instance when the LLM-generated summaries are misleading or too broad (\enquote{performed various actions}).
Further, both approaches sometimes struggle with reasoning or correctly using the tree exploration API (\eg getting stuck in repetitive behavior), which is most likely related to the difficulty of the questions and the general LLM capabilities.

\section{Conclusion \& Discussion}
We present \ours, a system for verbalization of life-long robot experience.
The multimodal, hierarchical representation of EM is interactively accessed by an LLM to answer user questions, keeping token costs low even for extremely long histories.
Despite the promising results and versatility of our system, it has some limitations:
First, as a modulated approach, it is limited by the performance of each component and can suffer from error propagation.
While the interactive tree search improves interpretability, there are no performance guarantees.
Moreover, our system could integrate more modalities and tools, \eg joint angle proprioception data could be rendered in simulation and then verbalized by a VLM.
Episodic memories should refer back to previous events to enable contextual or comparative descriptions \cite{perera_language-based_2018}.
Adding personalization, both to EM and verbalization, is desirable for improved human-robot interactions.
We hope our code and data will foster research on EMV, and will continue addressing these challenges in future work.

%% file: figures/tikz/tokens_vs_performance_per_ep_len.tikz
\begin{tikzpicture}[baseline]
\scriptsize
\begin{axis}[
    width=\linewidth,
    height=5.5cm,
    xlabel=1K Tokens per sample ($T$),
    xmode=log,
    ymin=0,
    ymax=100,
    enlarge x limits=0.07,
   	legend style={
        at={(1,0.99)}, anchor=north east, legend columns=2,
        cells={anchor=west},draw=none,fill=none
    },
    yticklabel={\pgfmathparse{\tick}\pgfmathprintnumber{\pgfmathresult}\%},
    ]
\legend{\ours 1-shot, \ours 0-shot, \ours (L3), text 1-pass L3}
\addplot[mark options={solid}, color=red,   mark=+,dashed,forget plot]  table[x=1s T,y=1s S_c,		col sep=tab, /pgf/number format/read comma as period]{figures/tikz/data.tsv};
\addplot[mark options={solid}, color=red,   mark=x,solid             ]  table[x=1s T,y=1s S_p,       col sep=tab, /pgf/number format/read comma as period]{figures/tikz/data.tsv};
\addplot[mark options={solid}, color=black, mark=+,dashed,forget plot]  table[x=0s T,y=0s S_c,       col sep=tab, /pgf/number format/read comma as period]{figures/tikz/data.tsv};
\addplot[mark options={solid}, color=black, mark=x,solid             ]  table[x=0s T,y=0s S_p,       col sep=tab, /pgf/number format/read comma as period]{figures/tikz/data.tsv};
\addplot[mark options={solid}, color=blue,  mark=+,dashed,forget plot]  table[x=predef T,y=predef S_c,	col sep=tab, /pgf/number format/read comma as period]{figures/tikz/data.tsv};
\addplot[mark options={solid}, color=blue,  mark=x,solid             ]  table[x=predef T,y=predef S_p,	col sep=tab, /pgf/number format/read comma as period]{figures/tikz/data.tsv};
\addplot[mark options={solid}, color=orange,mark=+,dashed,forget plot]  table[x=1p T,y=1p S_c,       col sep=tab, /pgf/number format/read comma as period]{figures/tikz/data.tsv};
\addplot[mark options={solid}, color=orange,mark=x,solid             ]  table[x=1p T,y=1p S_p,       col sep=tab, /pgf/number format/read comma as period]{figures/tikz/data.tsv};
\end{axis}
\end{tikzpicture}%

%% file: figures/tikz/teach-retrieval-f1-1shot.tikz
\begin{tikzpicture}[baseline]
\scriptsize
\begin{axis}[
    width=0.9\linewidth,
    height=3.5cm,
    ylabel=Retrieval $F_1$,
    xlabel=$|h|$,
    xlabel style={
        at={(ticklabel cs:1)},
        anchor=south west,
    },
    ymin=0,
    ymax=1,
    enlarge x limits=0.07,
   	legend style={
        at={(1,1)}, anchor=north east, legend columns=2,
        cells={anchor=west},draw=none
    },
    xtick=data,
    ]
\legend{overall, only correct, only part. corr., only wrong}
\addplot[mark options={solid}, color=black,   mark=+,dashed]  table[x=ep len,y=overall               , col sep=tab, /pgf/number format/read comma as period]{figures/tikz/data-teach-retrieval-f1-1shot.tsv};
\addplot[mark options={solid}, color=olive, mark=x,solid]  table[x=ep len,y=only_correct          , col sep=tab, /pgf/number format/read comma as period]{figures/tikz/data-teach-retrieval-f1-1shot.tsv};
\addplot[mark options={solid}, color=cyan,  mark=x,solid]  table[x=ep len,y=only_partially_correct,	col sep=tab, /pgf/number format/read comma as period]{figures/tikz/data-teach-retrieval-f1-1shot.tsv};
\addplot[mark options={solid}, color=red,mark=x,solid]  table[x=ep len,y=only_wrong            , col sep=tab, /pgf/number format/read comma as period]{figures/tikz/data-teach-retrieval-f1-1shot.tsv};
\end{axis}
\end{tikzpicture}%